\documentclass[journal]{IEEEtran}
\usepackage{cite}
\ifCLASSINFOpdf
   \usepackage[pdftex]{graphicx}
   \graphicspath{{../pdf/}{../jpeg/}}
   \DeclareGraphicsExtensions{.pdf,.jpeg,.png}
\else
   \usepackage[dvips]{graphicx}
   \graphicspath{{../eps/}}
   \DeclareGraphicsExtensions{.eps}
\fi
\usepackage{amsmath}
\usepackage{algorithmic}
\usepackage{array}
\ifCLASSOPTIONcompsoc
  \usepackage[caption=false,font=normalsize,labelfont=sf,textfont=sf]{subfig}
\else
  \usepackage[caption=false,font=footnotesize]{subfig}
\fi

\usepackage{stfloats}
\usepackage{url}
\usepackage{booktabs}       % professional-quality tables
\usepackage{amsfonts}       % blackboard math symbols
\usepackage{multirow}
\usepackage{amsmath}

\begin{document}
%\linenumbers	
\title{A Unified Light Framework for Real-time Fault Detection of Freight Train Images}
\author{Yang~Zhang,~Moyun~Liu,~Yang~Yang,~Yanwen~Guo,~and~Huiming~Zhang
%\thanks{Manuscript received ** **, 2020; revised ** **, 2020. accepted ** **, 2021.
%Date of publication ** **, 2021; date of current version ** **, 2021. This work was
%supported in part by the National Natural Science Foundation of China
%(Grant 62032011, 61772257, 51675166, and 61672279), and the program B for Outstanding PhD
%candidate  of  Nanjing University (Grant 202001B054). Paper no. TII-20-4527. 
%\emph{(Corresponding author: Moyun~Liu.)}}
\thanks{Y.~Zhang is with the School of Mechanical Engineering, Hubei University of 
Technology, Wuhan 430068, China, and also with the National Key
Laboratory for Novel Software Technology, Nanjing University, Nanjing 210023,
China (e-mail: yzhangcst@smail.nju.edu.cn)}
\thanks{M. Liu is with the School of Mechanical Science and Engineering,
Huazhong University of Science and Technology, Wuhan 430074, China
(e-mail: lmomoy8@gmail.com)}
\thanks{Y.~Yang, Y.~Guo, H.~Zhang and are with the National Key
Laboratory for Novel Software Technology, Nanjing University, Nanjing 210023,
China (e-mail: yyang\_nju@outlook.com; ywguo@nju.edu.cn; zhanghmcst@163.com).}% <-this % stops a space
%\thanks{Color versions of one or more of the figures in this article are 
%available online at http://ieeexplore.ieee.org.}
%\thanks{Digital Object Identifier}
}


\markboth{IEEE TRANSACTIONS ON INDUSTRIAL INFORMATICS,~Vol.~**, No.~*, ***~****}%
{Shell \MakeLowercase{\textit{et al.}}: Bare Demo of IEEEtran.cls for IEEE Journals}



\maketitle

\begin{abstract}
  Real-time fault detection for freight trains plays a vital role in guaranteeing the security and optimal operation of railway transportation under stringent resource requirements. Despite the promising results for deep learning based approaches, the performance of these fault detectors on freight train images, are far from satisfactory in both accuracy and efficiency. This paper proposes a unified light framework to improve detection accuracy while supporting a real-time operation with a low resource requirement. We firstly design a novel light-weight backbone  (RFDNet) to improve the accuracy and reduce computational cost. Then, we propose a multi region proposal network using multi-scale feature maps generated from RFDNet to improve the detection performance. Finally, we present multi level position-sensitive score maps and region of interest pooling to further improve accuracy with few redundant computations. Extensive experimental results on public benchmark datasets suggest that our RFDNet can significantly improve the performance of baseline network with higher accuracy and efficiency. Experiments on six fault datasets show that our method is capable of real-time detection at over 38 frames per second and achieves competitive accuracy and lower computation than the state-of-the-art detectors.
\end{abstract}

\begin{IEEEkeywords}
Real-time, fault detection, freight train images, light-weight, multi-scale.
\end{IEEEkeywords}

% \fi
\IEEEpeerreviewmaketitle

\section{Introduction}
\label{intro}
\begin{figure*}[!t]
 \centering
 \subfloat[]{
 \includegraphics[height=1.5in]{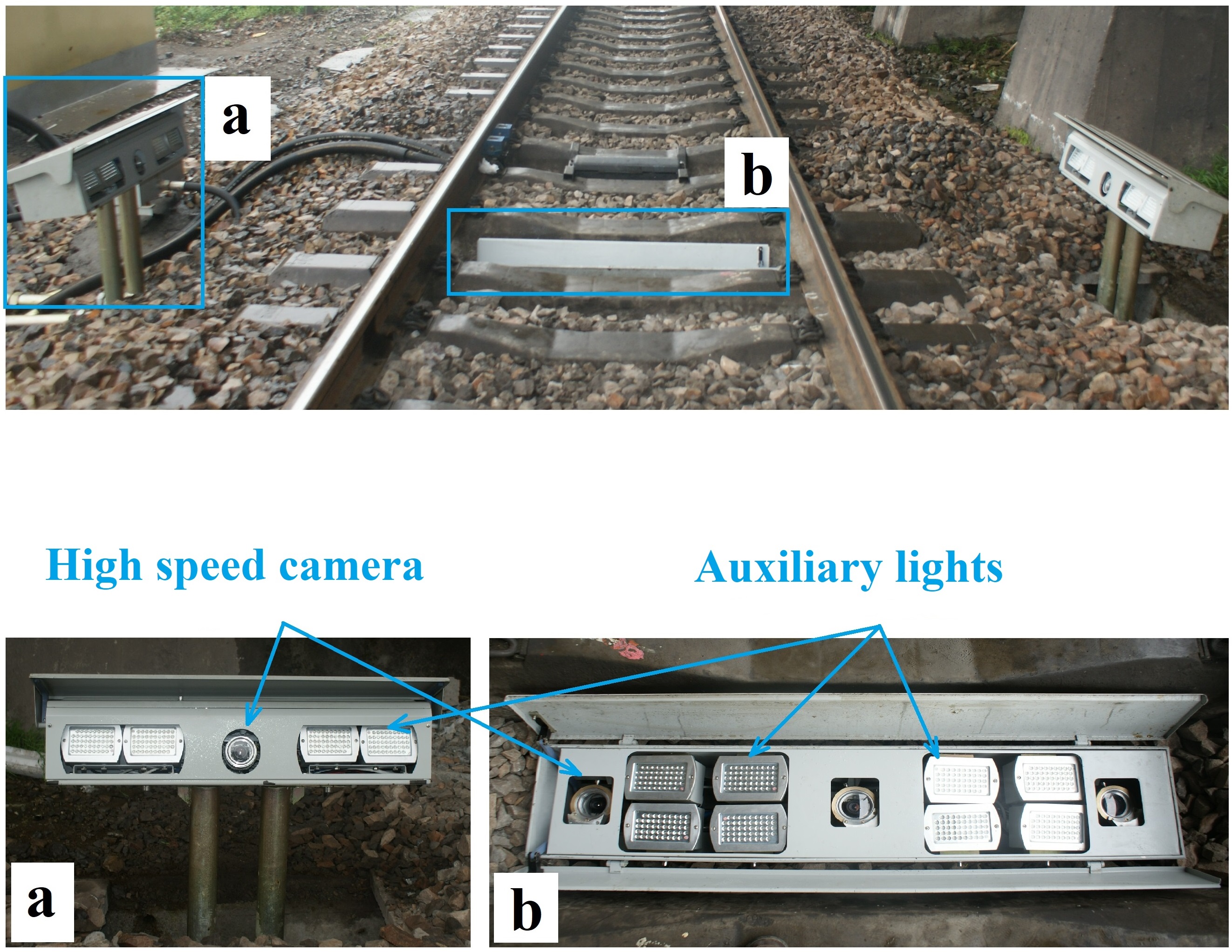}}
 \hspace{1em}
 \subfloat[]{
 \includegraphics[height=2.2in]{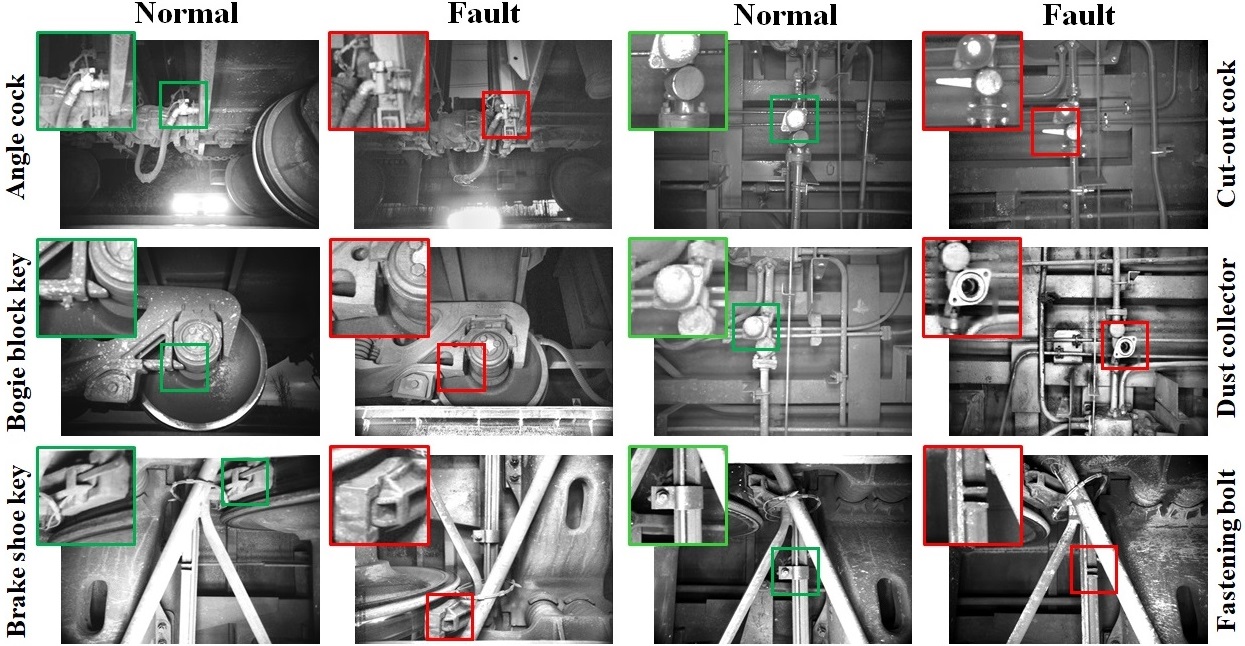}}
 \caption{Real-time fault detection for freight train images. (a) Image acquisition devices contain high speed cameras and auxiliary lights, which are installed on both sides and in the middle of railway tracks. (b) Some typical samples of freight train images. Some samples of freight train images are used to train fault detector. The final detection results include the location of the fault and its type.}
 \label{fig:detection}
\end{figure*}

\IEEEPARstart{F}{ault} detection is a vital routine maintenance work with regard to railway system~\cite{8673908,8692707,8667891}. For the freight trains, vehicle braking and steering systems contain many important parts that need to be carefully detected, because the loss or displacement of these components will seriously affect the driving safety. Such detection task applies the vision-based methods to replace the manual detection with many advantages such as high efficiency and accuracy. However, the image acquisition devices are installed outdoors as shown in Fig.~\ref{fig:detection}(a). The illumination variation always impacts the quality of acquired images as shown in Fig.~\ref{fig:detection}(b). It is difficult to possess sufficient features on account of that various parts are usually small, polluted, or obscure. These parts usually contain too much structural information, and the textures are similar to the backgrounds. All these problems always lead to failure in fault detection for freight train images. Moreover, the only resource-constrained devices are available in practical applications due to field environmental limitations.

In general, vision-based fault detection can be considered as a special type of object detection task in computer vision~\cite{8911418}. Recently, the rapid development in deep learning techniques can provide a robust solution for object detection, because deep networks especially the convolutional neural networks (CNNs) actually implement the functions of higher complexity. Deep learning-based object detection methods can detect different objects simultaneously with higher accuracy, even in complicated and changeable environments. To get better results, researchers have designed deeper, broader and more complex networks such as faster region-based CNN (Faster R-CNN)~\cite{RenHGS15}, and region-based fully convolutional network (R-FCN)~\cite{DaiLHS16}. However, these superior CNN-based detectors face many difficulties when they are applied into real-time fault detection. For example, Zhang~\emph{et al}.~\cite{zhang2018} proposed a unified framework for fault detection of freight train images (FTI-FDet) based on Faster R-CNN~\cite{RenHGS15}. But it is insufficient to achieve fast speed, and its model size is huge ($>$550 MB). Based on the above FTI-FDet, Zhang \emph{et al}.~\cite{8911418} proposed a specialized light fault detector (Light FTI-FDet) which pursues a balance between accuracy and speed. However, its model size is still over 89 MB. Such approaches~\cite{zhang2018,8911418} have been proved to be accurate enough to meet the actual needs, but the effectiveness of CNN-based detectors especially the illumination invariance, is missing to be analyzed in principle. More importantly, all previous studies~\cite{Sun2017Automatic,zhang2018,8911418,RenHGS15,DaiLHS16} have been unable to meet a real-time detection speed of above 30 frames per second (fps)~\cite{Gaussian-YOLOv3}, which is a prerequisite for vision-based fault detection. In addition, the model size of these detectors is also too large for strict memory and computational budget constraints. Therefore, a real-time fault detector for freight train images should first be robust to the illumination variation in the field environment. Then it needs to achieve an outstanding trade-off between speed and accuracy under stringent resource requirements which are not only computational cost for speed, but also memory resources on hardware.

To solve these problems, we propose a light and accurate framework to fulfill real-time fault detection task for freight train images. Over the years, many innovative light-weight networks have been proposed such as SqueezeNet~\cite{iandola2016squeezenet} and MobileNet~\cite{sandler2018inverted}, etc. Meanwhile, many researches are devoted to putting these light-weight backbones into practice~\cite{muhammad2019efficient, 8673908}. Inspired by the SqueezeNet, we firstly design a novel real-time fault detection network (RFDNet) as a backbone to improve accuracy while optimizing the network to meet resource requirement (hardware). It is also proved that our proposed RFDNet is robust to the illumination variation in freight train images. To improve the detection performance, we introduce a fault multi region proposal network (multi-RPN) by fusing the features from multiple layers in RFDNet. Unlike superior detectors such as Faster R-CNN and R-FCN only perform region of interest (RoI) pooling, we present multi level position-sensitive (MLPS) score maps and RoI pooling by using multi scale features for fault region detection. Experimental results on ImageNet, PASCAL VOC, and MS COCO datasets demonstrate that our RFDNet achieves much better performance than SqueezeNet. In addition, the extensive experiments on six fault datasets show that our framework can be effectively applied to achieve real-time fault detection at over 38 fps. Our framework achieves competitive accuracy and lower resource requirements such as 19.6 MB model size, compared with the state-of-the-art detectors.

In summary, this work makes the following contributions.
\begin{itemize}
  \item We design a light and accurate framework to achieve real-time fault detection for freight train images under stringent resource requirements.
  \item We propose a light-weight backbone RFDNet to significantly improve detection accuracy and reduce computational cost, which is confirmed to be robust to illumination changes.
  \item We introduce multi-RPN and MLPS by using multi-scale feature maps generated from RFDNet to improve the detection performance with few redundant computations.
  \item We validate the effectiveness of our RFDNet on public benchmark datasets and our method on six fault datasets with thorough ablation studies. Compared with the state-of-the-art methods, our framework achieves real-time detection at over 38 fps with competitive accuracy and lower resource requirements.
\end{itemize}

The rest of this paper is organized as follows. Section II presents some related works about fault detection for freight train images, object detection, and light-weight neural networks. Section III describes our framework and each important module. Comprehensive experiments are shown in Section IV to validate the superiority of our method and finally Section V concludes this paper.

\begin{figure*}[!t]
 \centering
 \includegraphics[width=6.3in]{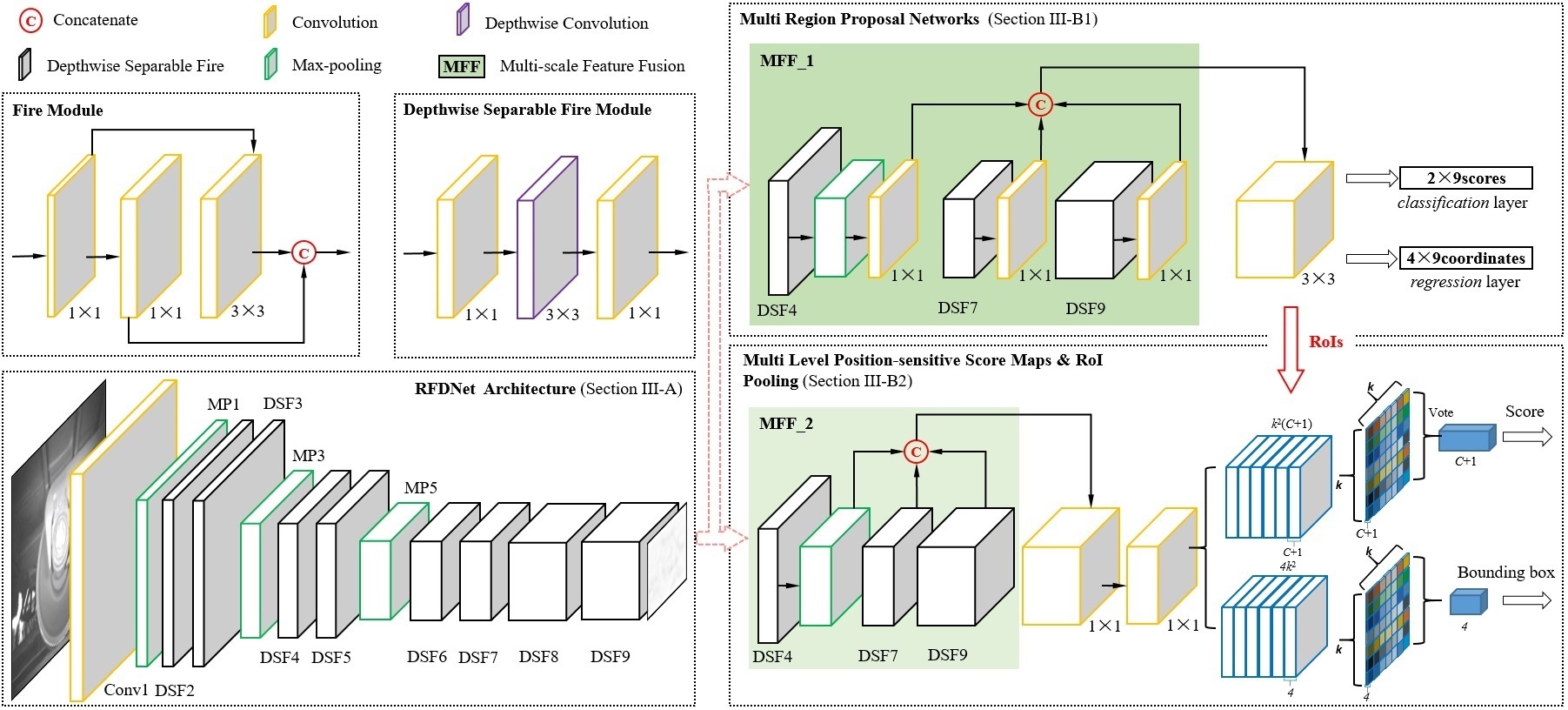}
 \caption{Pipeline of our proposed framework for real-time fault detection of freight train images. Our framework consists of three parts: real-time fault detection network (RFDNet), multi region proposal network, as well as multi level position-sensitive score maps and RoI pooling. The proposed framework takes an image as input, generates hundreds of fault region proposals via multi region proposal network from RFDNet, and then scores each proposal using multi level position-sensitive score maps and RoI pooling.}
 \label{fig:pipeline}
\end{figure*} 
\section{Related works}
\label{relatedworks}
\textbf{Fault Detection for Freight Train Images.}
Some of the recent researches for fault detection of freight train images are listed as follows. Liu \emph{et al}.~\cite{Liu2015Automated} proposed a hierarchical fault inspection framework to detect the missing of bogie block key on freight trains with high speed and accuracy. Zheng \emph{et al}.~\cite{Zheng2016Automatic} proposed an automatic image inspection system to inspect coupler yokes by a linear support vector machine (SVM) classifier for localization and Adaboost decision trees for recognition. However, these methods only detect one type of faults, which greatly influence their effectiveness. In addition, Sun \emph{et al}.~\cite{Sun2018Railway} proposed a fast adaptive Markov random field (FAMRF) for image segmentation and an exact height function (EHF) for shape matching of fault region. This method solves the problem of multi-fault detection, but it is too complex to achieve enough accuracy and fast speed. Differing from the conventional techniques, deep learning methods deal with more complex and difficult problems in machine vision field. Sun \emph{et al}.~\cite{Sun2017Automatic} presented a CNN-based system consisting of two complex models for target region detection and fault recognition, respectively. Pahwa \emph{et al}.~\cite{8917062} performed a two-step high resolution segmentation of the train valves and use image processing techniques to identify faulty valves. Fu \emph{et al}.~\cite{FU2020212} proposed a two-stage method cascading a bearing localization stage and a defection segmentation stage to recognize the defect areas in a coarse-to-fine manner. However, these methods have high computational cost, which are insufficient to meet actual requirements of fault detection like real-time and versatility.

\textbf{Object Detection.}
As the basis of fault detection, object detectors based on CNNs have been developed rapidly over the years, which are widely used in actual applications due to their powerful capacity~\cite{lu2018estimation}. These CNN-based detectors can be divided into two parts: one-stage and two-stage. One-stage detectors directly predict object classes and locations without region proposal generation, such as you only look once (YOLO)~\cite{redmon2018yolov3}, single shot multibox detector (SSD)~\cite{LiuAESRFB16}, reverse connection with objectness prior network (RON)~\cite{kong2017ron}, RefineDet~\cite{zhang2018single}, and deeply supervised object detector (DSOD)~\cite{shen2017dsod}. Based on one-stage strategy, these detectors can obtain fast speed, which are suited for limited computing condition, but usually sacrifices accuracy. Two-stage detectors firstly generate a set of region proposals, and then classify whether they are background or foreground. These detectors achieve accurate and effective object detection such as Faster R-CNN~\cite{RenHGS15}, R-FCN~\cite{DaiLHS16}, multi-scale location-aware kernel representation (MLKP)~\cite{wang2018multi}, and Cascade R-CNN~\cite{cai2018cascade}. Compared with one-stage detectors, these two-stage detectors have higher performance but need more computations. Hence, they are usually incapable of coping with practical application for real-time demand.

\textbf{Light-weight Neural Networks.}
During past years, many efforts are devoted to designing light-weight backbones for object detection task in the resource-restricted conditions. There are some light-weight architecture designs which achieve better speed-accuracy trade-offs, including SqueezeNet~\cite{iandola2016squeezenet}, MobileNet~\cite{sandler2018inverted}, and ShuffleNet~\cite{ma2018shufflenet}, etc. Compared with the superior performance models like ResNet~\cite{He2016Deep}, these networks have fewer parameters with approximate precision. Moreover, there has been a growing interest in incorporating light-weight networks into CNNs for object detection task. For example, Tiny-DSOD~\cite{yuxi2018tinydsod} consists of depthwise dense block based backbone and depthwise feature pyramid network, achieving a better trade-off between resources and accuracy. Pelee~\cite{wang2018pelee} combines a PeleeNet with SSD to keep detection accuracy for mobile applications with fast speed. Each of these light-weight neural networks has a small model size, but its accuracy still has a large room for improvement.

As for real-time fault detection task, both accuracy and computation complexity are important considerations under stringent resource requirements in field environment. To obtain an outstanding balance between accuracy and computational cost, we take inspiration from both the incredible efficiency of the Fire modules introduced in SqueezeNet~\cite{iandola2016squeezenet} and the powerful detection performance demonstrated by the R-FCN~\cite{DaiLHS16}. In addition, the multi-level feature fusion strategy in~\cite{8911418} can be used to further improve the detection accuracy notably without a lot redundant computations.

\section{The Proposed Framework}
\label{method}
In this section, we first introduce our proposed light-weight backbone RFDNet in Section~\ref{sec:DS-SqueezeNet}. To enrich features, we present a multi-RPN to combine different level feature maps, as introduced in Section~\ref{MRPN}. Then, we propose MLPS score maps and RoI pooling to better use the multi-level feature maps, as introduced in Section~\ref{MLPS}. The detailed architecture of the proposed framework is depicted in Fig.~\ref{fig:pipeline}. The proposed framework takes an image as input, generates hundreds of fault region proposals via multi-RPN from RFDNet, and then scores each proposal using MLPS score maps and RoI pooling.

\subsection{Light-weight Backbone}
\label{sec:DS-SqueezeNet}
In original R-FCN, the backbone (\textit{i.e.} ResNet~\cite{He2016Deep}) needs a large number of parameters and floating point operations (FLOPs) to achieve a satisfactory accuracy, thus requiring a huge amount of computations. However, as discussed in the previous section, fault detection task cannot meet the demand of tremendous computing power in the wild. Compared to the ResNet, a light-weight network usually has fewer parameters and lower computations with approximate precision, which is more suitable for real-time fault detection. SqueezeNet~\cite{iandola2016squeezenet} is an efficient network which uses a bottleneck approach to design a small-size network. Nevertheless, there is still a large accuracy gap between these networks and those of full-sized counterparts for detection~\cite{yuxi2018tinydsod}. For traditional SqueezeNet, it is a challenge to increase accuracy while reducing computing cost. In Fig.~\ref{fig:pipeline}, the core of SqueezeNet is Fire module which consists of a squeeze layer and an expand layer. The expand layer contains two layers: 3$\times$3 convolutional layer and 1$\times$1 convolutional layer. However, it is unreliable to preset the kernel numbers of 1$\times$1 and 3$\times$3 convolutional layers in expand layers. As an alternative, we remove the 1$\times$1 convolutional and retain the 3$\times$3 convolutional, and the network is adjusted to be a streamline.
In general, we define a $K_c\times K_c\times P\times Q$ convolutional (Conv.) kernel $K(\cdot)$, where $K_c$ is the spatial dimension of the kernel assumed to be square. \textit{P} is the number of input channels, and \textit{Q} is the number of output channels. The kernel $K(\cdot)$ slides on an input feature map $F(\cdot)$ to extract output features maps $G(\cdot)$ as follows~\cite{sandler2018inverted}:
\begin{equation}
G_{m, n, q}=\sum_{i,j,p}K_{i, j, p, q}\cdot F_{m+i-1, n+j-1, p} \;.
\end{equation}

\begin{table}[!t]
\renewcommand{\arraystretch}{1.03}
\centering
\caption{RFDNet detailed architecture}
\begin{tabular}{l|l|l}
\toprule
Layer name & Type / Stride & Filter shape \\
\midrule
Conv 1                                                   &Conv / s2  &3$\times$3$\times$3$\times$64   \\
\hline
MP 1                                             &MaxPooling / s2  &Pool 3$\times$3    \\
\hline
\multirow{3}{*}{{\begin{tabular}[l]{@{}c@{}} DSF 2 / 3 \end{tabular}}}    &Conv / s1   & 1$\times$1$\times$64$\times$64   \\ \cline{2-3}
           &Dw-Conv / s1 & 3$\times$3$\times$64 dw         \\ \cline{2-3}
           &Conv / s1    & 1$\times$1$\times$64$\times$64  \\
\hline
MP 3                                             &MaxPooling / s2  & Pool 3$\times$3    \\
\hline
\multirow{3}{*}{{\begin{tabular}[l]{@{}c@{}} DSF 4 \end{tabular}}}    &Conv / s1    & 1$\times$1$\times$64$\times$128  \\ \cline{2-3}
           &Dw-Conv / s1 & 3$\times$3$\times$128 dw         \\ \cline{2-3}
           &Conv / s1  & 1$\times$1$\times$128$\times$128   \\
\hline
\multirow{3}{*}{{\begin{tabular}[l]{@{}c@{}} DSF 5 \end{tabular}}}    &Conv / s1    & 1$\times$1$\times$128$\times$128  \\ \cline{2-3}
           &Dw-Conv / s1 & 3$\times$3$\times$128 dw         \\ \cline{2-3}
           &Conv / s1  & 1$\times$1$\times$128$\times$128   \\
\hline
MP 5                                             &MaxPooling / s2  & Pool 3$\times$3    \\
\hline
\multirow{3}{*}{{\begin{tabular}[l]{@{}c@{}} DSF 6 \end{tabular}}} &Conv / s1  & 1$\times$1$\times$128$\times$256      \\ \cline{2-3}
           &Dw-Conv / s1 & 3$\times$3$\times$256 dw         \\ \cline{2-3}
           &Conv / s1  & 1$\times$1$\times$256$\times$256   \\
\hline
\multirow{3}{*}{{\begin{tabular}[l]{@{}c@{}} DSF 7 \end{tabular}}} &Conv / s1  & 1$\times$1$\times$256$\times$256      \\ \cline{2-3}
           &Dw-Conv / s1 & 3$\times$3$\times$256 dw         \\ \cline{2-3}
           &Conv / s1  & 1$\times$1$\times$256$\times$256   \\
\hline
\multirow{3}{*}{{\begin{tabular}[l]{@{}c@{}} DSF 8 \end{tabular}}}  &Conv / s1  & 1$\times$1$\times$256$\times$512     \\ \cline{2-3}
           &Dw-Conv / s1 & 3$\times$3$\times$512 dw         \\ \cline{2-3}
           &Conv / s1  & 1$\times$1$\times$512$\times$512   \\
\hline
\multirow{3}{*}{{\begin{tabular}[l]{@{}c@{}} DSF 9 \end{tabular}}}  &Conv / s1  & 1$\times$1$\times$512$\times$512     \\ \cline{2-3}
           &Dw-Conv / s1 & 3$\times$3$\times$512 dw         \\ \cline{2-3}
           &Conv / s1  & 1$\times$1$\times$512$\times$512   \\
\hline
Conv 10                                                  &Conv / s1  & 1$\times$1$\times$512$\times$1000   \\
\hline
Avgpooling                                              &Average Pooling  / s1  & Pool 14$\times$14  \\
\hline
SoftmaxWithLoss                                         &Softmax / s1  & Classifier  \\
\bottomrule
\end{tabular}
\label{Dsqunetarchitecture}
\end{table}

Moreover, depthwise separable Conv.~\cite{sandler2018inverted} has shown computing efficiency in generic image classification tasks, drastically reducing computational cost and model size. In RFDNet, we use depthwise separable Conv. to improve the performance of Fire module, called as depthwise separable Fire (DSF) module. The proposed DSF contains a depthwise Conv. (Dw-Conv) and a pointwise Conv. layer. We use 3$\times$3 Dw-Conv to replace original expand layers of each Fire module in SqueezeNet. The Dw-Conv~\cite{sandler2018inverted} is defined as:
\begin{equation}
\hat{G}_{m, n, p}=\sum_{i,j} \hat{K}_{i, j, p}\cdot F_{m+i-1, n+j-1, p} \;,
\end{equation}
where $\hat{K}(\cdot)$ is the Dw-Conv kernel of size $K_c\times K_c\times P$, and $\hat{G}(\cdot)$ is the filtered output feature map. Such an approach achieves 8$\times$ less computation than standard Conv.~\cite{sandler2018inverted}. Pointwise Conv., namely a simple 1$\times$1 Conv., is then applied to create a linear combination of the output of depthwise layer. Both batch normalization (BN)~\cite{Ioffe2015Batch} and rectified linear unit (ReLU) nonlinearities are used for all layers in the DSF. The detailed architecture can be found in Table~\ref{Dsqunetarchitecture}. The RFDNet\footnote{For real-time fault detection, we remove the average pooling and the final layer, and only use the Conv.1 layer and DSF2$\sim$9 as the backbone.} begins with a standalone Conv. layer (Conv1) and eight DSF modules (DSF2$\sim$9), followed by a global average pooling, ending with a 1000-d 1$\times$1 Conv. layer.

\begin{figure}[!t]
	\centering
    \includegraphics[width=3.3in]{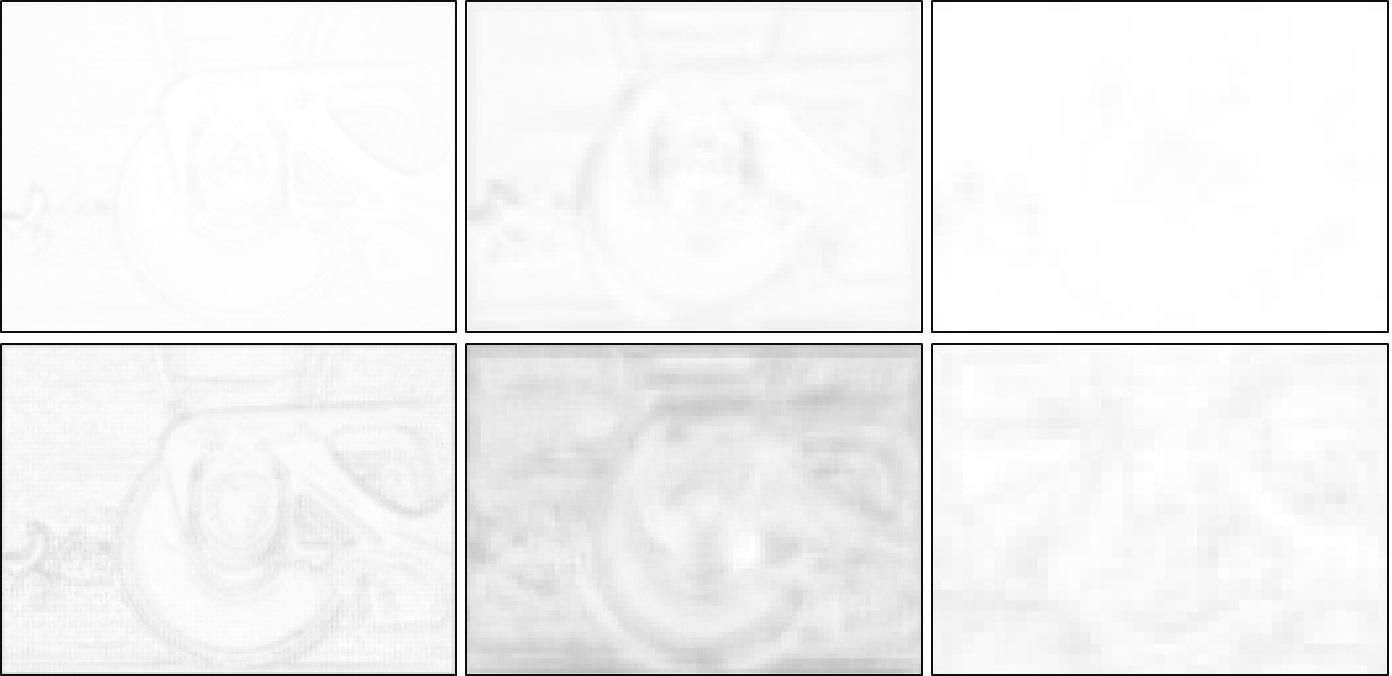}
	\caption{Visualization comparison of average feature maps extracted by Fire modules in SqueezeNet (top) and by the corresponding DSF modules in RFDNet (bottom). The average feature maps over all channels from Fire3, Fire5, and Fire7 are shown from left to right, respectively. The average feature maps over all channels from DSF3, DSF5, and DSF7 are shown from left to right, respectively. DSF modules remain more textural property in the low-level layer which present better fine-grained object details. DSF modules also capture more semantic cues in the high-level layer.}
	\label{fig:RFDNet_vis}
\end{figure}

To verify the advantage of our DSF intuitively, we calculate the average feature maps over all channels from Fire3, Fire5, and Fire7 in SqueezeNet, and the corresponding DSF3, DSF5, and DSF7 in our RFDNet, respectively. The visualization comparisons of extracted feature maps are demonstrated in Fig.~\ref{fig:RFDNet_vis}. The results indicate that DSF produces more salient features, while Fire misses some valuable information. In addition, the feature maps from DSF remain more textural property in the low-level layer. In this case, our DSF can show better fine-grained object details. Furthermore, our proposed DSF can capture more semantic cues in the high-level layer. The effectiveness of our DSF in RFDNet will be further described in the following Section~\ref{analysisframework}.

\begin{figure}[!t]
 \centering
 \subfloat[]{
 \includegraphics[width=1.05in]{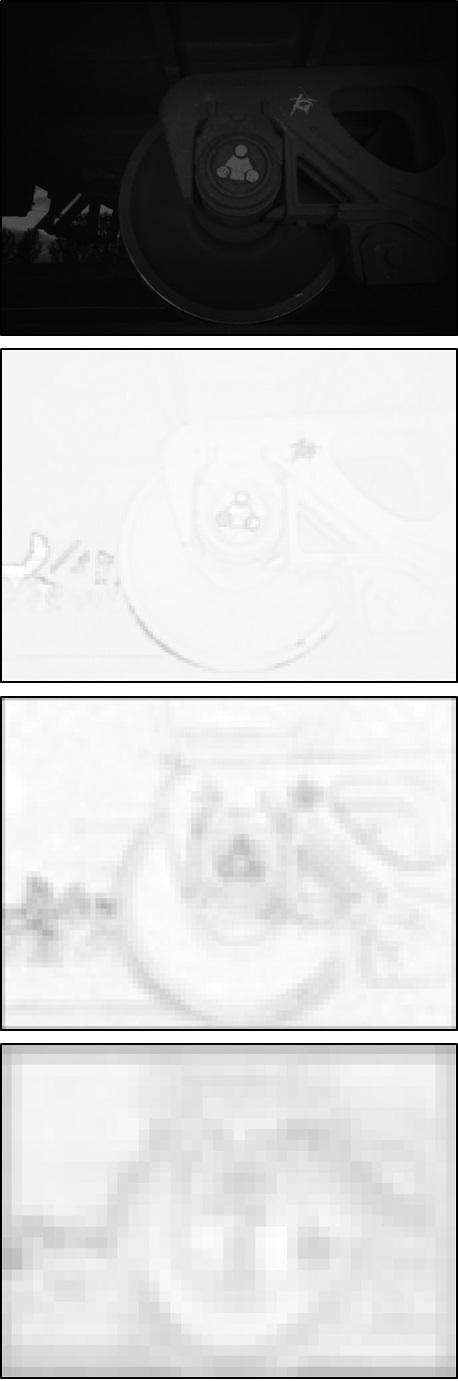}}
 \subfloat[]{
 \includegraphics[width=1.05in]{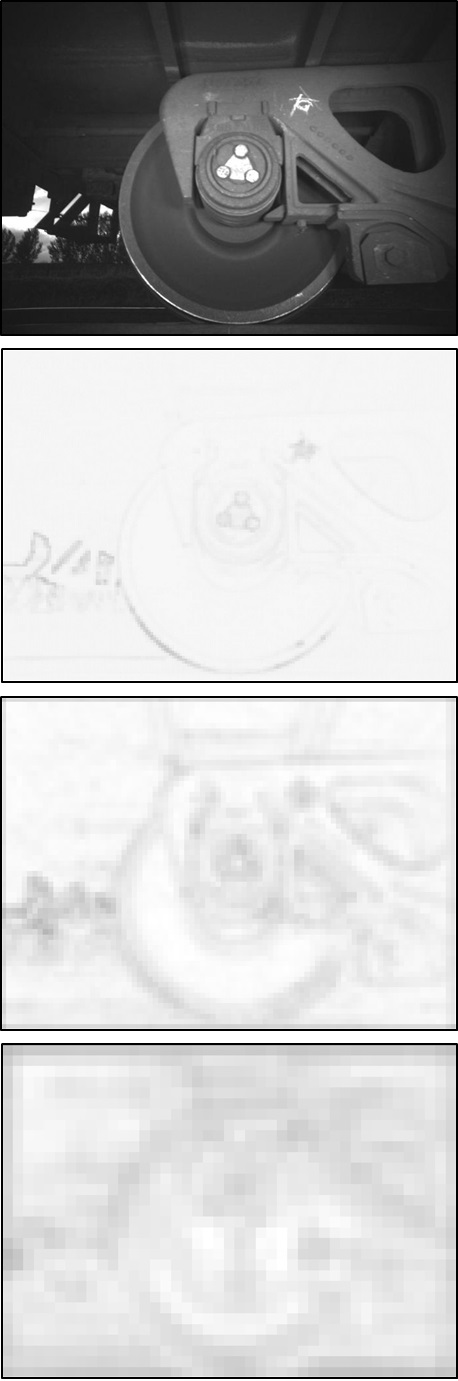}}
 \subfloat[]{
 \includegraphics[width=1.05in]{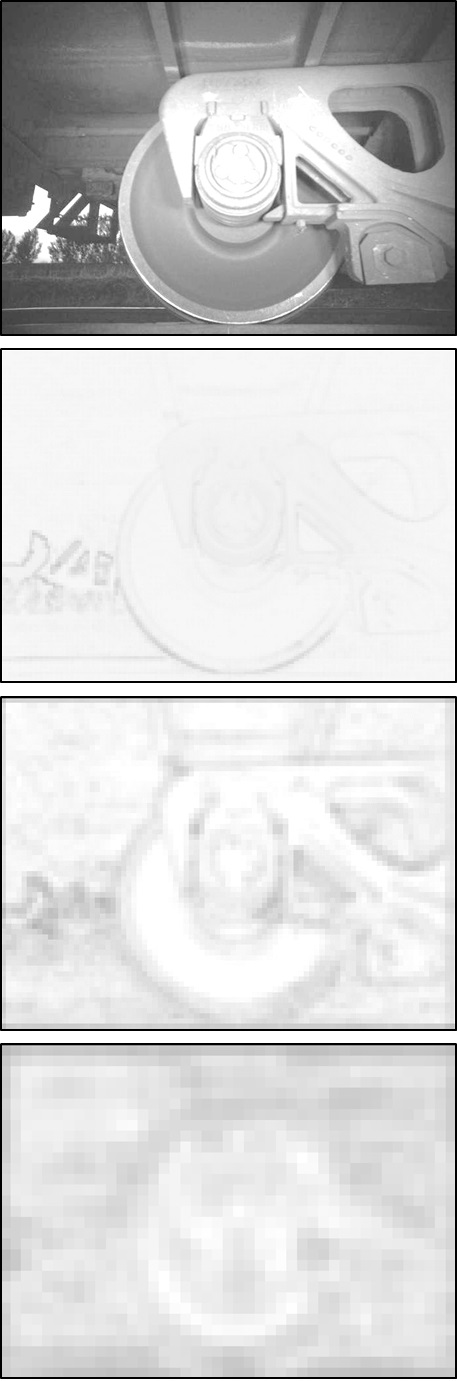}}
 \caption{Illustration of RFDNet for freight train images in varying light conditions. (a) Input image in low illumination with average feature maps of MP1, MP3 and MP5 from RFDNet. (b) Normal illumination image with the corresponding average feature maps. (c) High illumination image with the corresponding average feature maps. The feature maps derived from different inputs are similar at the same stage of networks, which means that RFDNet is robust to the change of illumination for fault detection.}
 \label{fig:conv}
\end{figure}

Besides the rich salient features, our proposed RFDNet can generate feature description which is invariant to illumination. Aiming at the main variation in illumination for freight train images, its robustness can be proved by feature maps extracted from different layers in this paper. In Fig.~\ref{fig:conv}, we firstly obtain the images under different illumination intensity. The feature maps are then extracted from different intermediate layers (MP1, MP3 and MP5) of RFDNet. We can observe that feature maps derived from different inputs are similar at the same stage of networks. This means that RFDNet is robust to the change of illumination for freight train images. We attribute this success to abundant input data and self-learning capacity of RFDNet, which can automatically learn to obtain better feature maps. Therefore, the introduction of RFDNet is advisable, because the weather and sunlight would make a great difference in light intensity for freight train images, which is common in practice. %Moreover, some modules in CNNs such as batch normalization (BN)~\cite{Ioffe2015Batch} can enhance generalization ability of network by normalizing samples to a certain distribution range.

\subsection{Multi-scale Feature Utilization}
Lower-level to higher-level layers in CNNs usually possess diverse distinguishing features for different size of objects. It can be seen from Figs.~\ref{fig:RFDNet_vis} and~\ref{fig:conv} that lower-level layers with higher resolution can capture more fine-grained information, which is helpful for recognizing small objects. Higher-level layers are more sensitive to semantic cues than lower-level layers. Therefore, multi-scale features can better represent all objects by incorporating multiple spatial resolutions in images. For freight train images, the detected parts have a range in size so that a single feature map cannot support for a satisfactory detection performance. So, we apply a multi-scale feature to produce more powerful feature maps of fault region, which can help to detect different size of objects.

\subsubsection{Multi-RPN}
\label{MRPN}
The function of RPN is to quickly select some candidate regions for target objects, which can greatly decrease the computation burden for inference process. A set of rectangular object proposals are usually generated by a fully Conv. network on feature maps. How to build an accurate RPN is important for two-stage detectors, and one potential way to improve its performance is employing multi-scale features.

We propose a novel RPN using a multi-scale feature fusion (MFF) block (see details of MFF\_1 in Fig.~\ref{fig:pipeline}) to apply multi-scale sliding windows over multi-level DSFs, which associates a set of prior anchors with each sliding position to generate fault region proposals. Specially, according to the size of fault regions, we use a 3$\times$3 sliding window who carries 9 anchors with 3 scales and 3 aspect ratios over the MFF\_1 block to produce multiple spatial features. To adjust multi-level feature maps to the same resolution for combination, different DSFs are processed by different sampling strategies. For DSF4, a 2$\times$2 max pooling layer is added to carry out subsampling. Then, we use 192-d 1$\times$1 Conv. to extract local feature over the above processed DSF4, DSF7, and DSF9, respectively. We normalize multiple feature maps using BN and then concatenate them. We encode the above concatenated feature maps using a 512-d 3$\times$3 Conv. layer which not only extracts more semantic features but also compresses them into a uniform space. The 512-d feature is then entered into two output layers: a classification layer that predicts the score of fault region, and a regression layer that refines the location for each prior anchor.
We define a bounding box as $t=(t_x,t_y,t_w,t_h)$ with the score $s$, and our loss function defined on each RoI is the summation of cross-entropy loss $L_{cls}$ and box regression loss $L_{reg}$~\cite{RenHGS15}:
\begin{equation}
\begin{split}
L(s,t_{x,y,w,h})&=L_{cls}(s_{c^*})+\lambda[c^*>0]L_{reg}(t,t^*) \\
&=-log(s_{c^*})+\lambda[c^*>0]L_{reg}(t,t^*),
\end{split}
\end{equation}
where $c^*$ denotes the ground-truth label of a RoI, and $t^*$ is the ground-truth bounding box. $\lambda$ is a balance weight which is set as 1. $[c^*>0]$ is an indicator that equals to 1 if the argument is true and 0 otherwise. Besides that, all local features are pre-computed before multi-RPN and detection without redundant computation~\cite{8911418}. The effectiveness of multi-RPN will be further described in the following Section~\ref{analysisframework}. %In addition, we apply a linear nonmaximum suppression (NMS)~\cite{BodlaSCD17} with a threshold 0.7 to rapidly suppress the lower score boxes and retain the highest scoring anchor in the neighborhood.

\subsubsection{MLPS}
\label{MLPS}
To better use the multi-level features and enrich the different information of each anchor, we perform position-sensitive RoI pooling over MLPS score maps. Before encoding position information into each RoI, we use another MFF block (see details of MFF\_2 in Fig.~\ref{fig:pipeline}) and encode the concatenated feature with a 512-d 1$\times$1 Conv. layer to combine the multi-level features. We then attach a 256-d 1$\times$1 Conv. layer for reducing dimension.
After that, the multi-level weighted fusion feature is accessed to produce $k^2$ position-sensitive score maps for each of the $C$ categories ($k$ is set to 7 in practice~\cite{RenHGS15}), correspondingly all RoIs also are evenly divided into $k^2$ grid areas. The MLPS scores vote on the RoI by averaging the scores, which is MLPS RoI pooling that can be denoted as:
\begin{equation}
P_{c}|\mathcal{A}_{(m,n)}=\sum_{i=1}^{N} \frac{1}{N}p_{(i)}^{(c)}|L_{m,n,c} \quad p\in \mathcal{A}_{(m,n)},
\end{equation}
where the $\mathcal{A}_{(m,n)}$ is an area within $k^2$ grids in each RoI, and it represents the location for specified area (0$\leq$ $m,n$ $\leq$  $k-1$). $P_{c}|\mathcal{A}_{(m,n)}$ is the pooling result for category $C$ at $\mathcal{A}_{(m,n)}$, and $p$ is the pixel in $\mathcal{A}_{(m,n)}$. $N$ denotes total number of pixels in $\mathcal{A}_{(m,n)}$, while $L_{m,n,c}$ is one of score maps that corresponds to $\mathcal{A}_{(m,n)}$ in $k^2$ score maps for $C$.

Finally, a ($C+1$)-d vector is produced for classification, and an average vote is used over the vector as follows
\begin{equation}
P_{c}= \sum _{m,n} P_{c}|\mathcal{A}_{(m,n)},
\end{equation}
where $P_{c}$ is the final score for category $C$, and we then calculate the softmax responses across categories:
\begin{equation}
s_c=\frac{e^{P_{c}}}{\sum_{c'=0}^{C} e^{P_{c'}}}.
\end{equation}
These are used for computing the cross-entropy loss $L_{cls}$ during training and for ranking the RoIs during inference.

\begin{figure}[!t]
	\centering
    \subfloat[]{\includegraphics[width=1.05in]{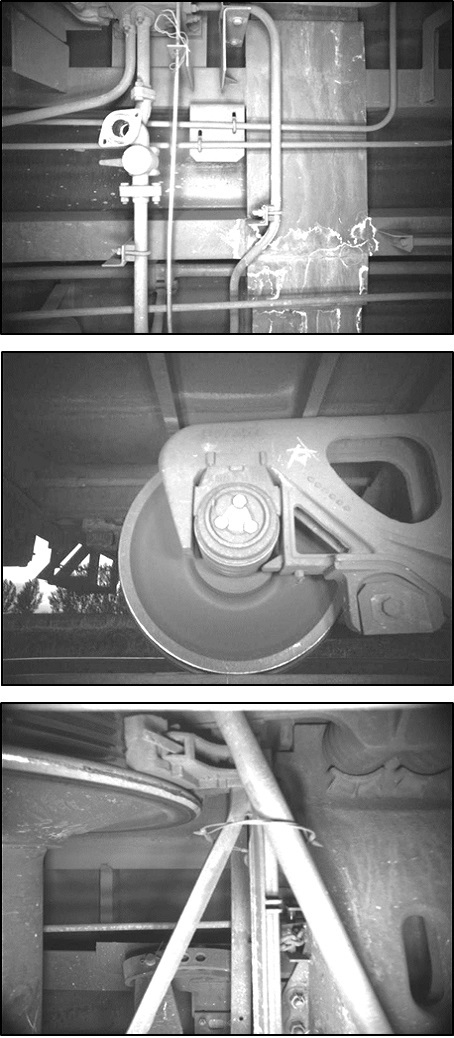}}\hspace{0.1em}
    \subfloat[]{\includegraphics[width=1.05in]{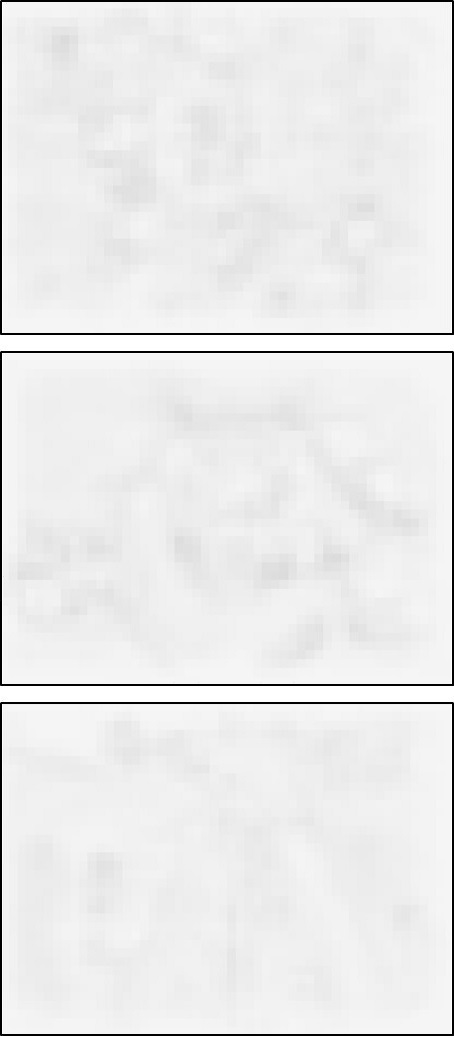}}\hspace{0.1em}
    \subfloat[]{\includegraphics[width=1.05in]{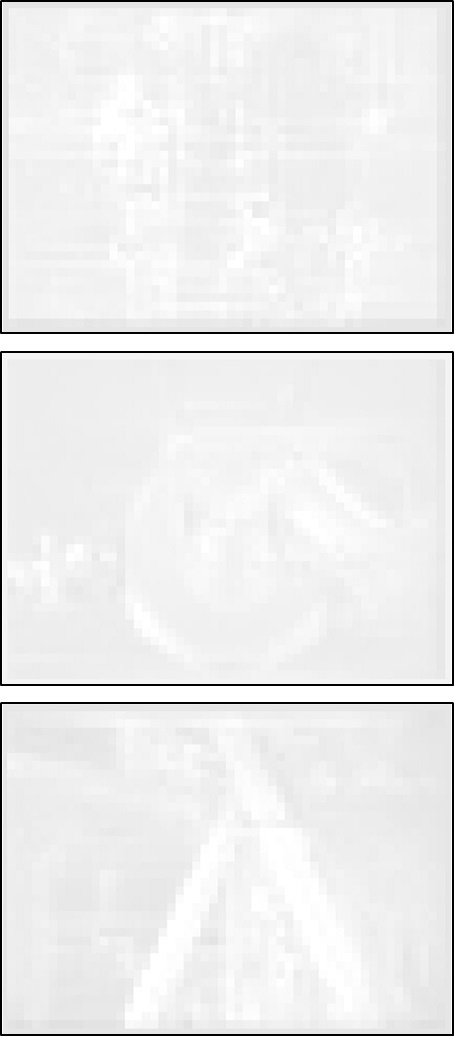}}
	\caption{Visualization results of the multi-level feature concatenation. (a) Input images; (b) Average feature maps extracted by DSF9; (c) Average feature maps concatenated by DSF4, DSF7, and DSF9. The multi-level fusion feature has richer object characteristic than a single, such as shape and contour.}
	\label{fig:vis_MLPS}
\end{figure}

Aiming at achieving bounding box regression, a sibling 4$k^2$-d Conv. layer is then appended for bounding box regression. The MLPS RoI pooling is performed on this bank of 4$k^2$ maps as well. Then, it is aggregated into a 4-d vector by average voting which is used to parameterize a bounding box. There is no learnable layer after the RoI, enabling nearly cost-free region-wise computation and speeding up both training and inference~\cite{RenHGS15}. The visualization results of the multi-level feature concatenation are demonstrated in Fig.~\ref{fig:vis_MLPS}. The average feature maps of DSF9 are extremely scarce for different freight train images, which only contain semantic cues with low resolution. The MLPS score maps and RoI Pooling will be unreliable to detect faults only based on the feature maps processed by DSF9. Nevertheless, the multi-level fusion feature has rich object characteristic such as shape and contour, which is helpful to improve the detection accuracy. The applicability of MLPS score maps and RoI Pooling will be further described in the following Section~\ref{analysisframework}.

\section{Experiments and Analysis}
\label{experiment}
In this section, we evaluate the effectiveness of our framework on the problem of real-time fault detection for freight train images. To this end, we firstly evaluate our proposed light-weight backbone on three datasets, ImageNet ILSVRC 2012~\cite{SimonyanZ14a}, PASCAL visual object classes (VOC) 2007~\cite{EveringhamEGWWZ15} and MS COCO~\cite{2014Microsoft}. Then we compare the proposed framework with state-of-the-art fault detectors and well-known object detection methods on six fault datasets~\cite{zhang2018,8911418}. We conduct all of our experiments using Caffe~\cite{JiaSDKLGGD14} on a single NVIDIA GeForce GTX1080Ti GPU.
%\begin{table}[!t]
%\renewcommand{\arraystretch}{1.1}
%\centering
%\caption{The datasets of fault detection for freight train images}
%\begin{tabular}{lcccc}
%\toprule
%\multirow{2}{*}{Datasets}  & \multirow{2}{*}{Training set} & \multicolumn{3}{c}{Testing set}
%\\ \cline{3-5}
%                    &              & Non-fault   & Fault & Total \\
%\midrule
%Angle cock          & 2002         & 1049        & 975   & 2024  \\
%Bogie block key     & 5440         & 2530        & 367   & 2897  \\
%Brake shoe key      & 5636         & 2000        & 2000  & 4000  \\
%Cut-out cock        & 815          & 671         & 179   & 850   \\
%Dust collector      & 815          & 798         & 52    & 850   \\
%Fastening bolt      & 1724         & 445         & 1257  & 1902  \\
%\bottomrule
%\end{tabular}
%\label{database}
%\end{table}

\subsection{Experimental Setup}
\subsubsection{Implementation Details}
In ImageNet experiments, to make a fair comparison, all the hyper-parameters follow SqueezeNet~\cite{iandola2016squeezenet}. We use BN after each Conv. layer before ReLU activation, and the initial learning rate is set to 0.04. We use the polynomial decay learning rate scheduling strategy in the batch size of 32. The momentum and weight decay are set as 0.9 and 0.0002, respectively. Finally, we use the validation set of ImageNet ILSVRC 2012 to validate our backbone.

In PASCAL VOC experiments, we use the same hyper-parameters as SqueezeNet to make a fair comparison. Based on a pre-trained model from the ImageNet experiments, we fine-tune the resulting model using RMSProp with 0.0001 initial learning rate, 0.9 momentum, and 0.0005 weight decay. We set 120K training steps and execute multi-scale training in the batch size of 64. We use the step decay learning rate scheduling strategy and multiply with a factor 0.1 at the 20K, 50K, and 100K steps, respectively. Finally, the VOC 2007 test set is used to verify our RFDNet following the protocol in~\cite{RenHGS15}.

In MS COCO experiments, we also use the same hyper-parameters as SqueezeNet for fair comparison. Based on a pre-trained model from the ImageNet experiments, we fine-tune the resulting model using SGD with 0.001 initial learning rate, 0.9 momentum, and 0.0005 weight decay. We set 480K training steps and execute multi-scale training in the batch size of 56. We use the step decay learning rate scheduling strategy and multiply with a factor 0.1 at the 280K, and 360K steps, respectively. Finally, the COCO minival set is adopted to evaluate our backbone following the standard protocol.

In fault detection experiments, our method is trained via back-propagation and stochastic gradient descent (SGD). We use a basic learning rate of 0.001 and it is divided by 10 for each 40K mini-batch until convergence. The batch sizes of  multi-RPN and MLPS RoI are 256 and 512, respectively. A pre-trained RFDNet model for ImageNet is first used to initialize shared Conv. layers of our backbone network, and then the new layers are initialized with a zero mean and a standard deviation of 0.01 Gaussian distribution. We train the network with 70K iterations in total. The momentum and weight decay are set as 0.9 and 0.0005, respectively. The confidence score in the detecting stage is 0.9.

\subsubsection{Fault Datasets} To evaluate the performance of our method, six fault datasets~\cite{zhang2018,8911418} for freight train images are directly used in this study, including angle cock, bogie block key, brake shoe key, cut-out cock, dust collector, and fastening bolt on brake beam. Some typical samples of freight train images are shown in Fig~\ref{fig:detection}(b).
\begin{itemize}
  \item \textbf{Angle cock} is a key component of the air brake system of freight trains, and its role is to ensure the smooth flow of air in the main pipeline. For this dataset, training and evaluation are performed on the 2002 images in the trainval and the 2024 images in the test, respectively.
  \item \textbf{Bogie block key} is a very small part used to prevent the wheel set from getting out of the bogie. This dataset is divided into two sets, training and testing with 5440 and 2897 images, respectively.
  \item \textbf{Brake shoe key} is also a small component equipped in brake shoe, which is vital for safe operation of braking system. The dataset provides more than 5600 images for training, and 4000 images for its test set.
  \item \textbf{Cut-out cock} is a key part that cuts off the air from main reservoir to the brake pipe, which is used to shut down the brake pipe. The images are divided into a train set of 815 images and a test set of 850 images.
  \item \textbf{Dust collector} is usually installed next to the cut-out cock and its role is to filter impurities towards compressed air. So, the images in this dataset are annotated directly on the images in cut-out cock dataset.
  \item\textbf{Fastening bolt} is an important part for train brake. When the train brakes, the fastening bolts may break or fall off because of a large horizontal force generated from brake beam. There are 1724 images in the train set and another 1902 images in the test set.
\end{itemize}

\begin{table}[!t]
\renewcommand{\arraystretch}{1}
\centering
\caption{Classification results on ImageNet ILSVRC 2012}
\begin{tabular}{lcccc}
\toprule
 \multirow{2}{*}{{\begin{tabular}[l]{@{}c@{}} Model \end{tabular}}}  & \multirow{2}{*}{{\begin{tabular}[c]{@{}c@{}} Computational \\ cost (FLOPs) \end{tabular}}}
& \multirow{2}{*}{{\begin{tabular}[c]{@{}c@{}} Model size\\(Parameters)\end{tabular}}} & \multirow{2}{*}{Top-1} & \multirow{2}{*}{Top-5}\\
&&&&\\
\midrule
SqueezeNet                   & 833M  & 4.8MB  & 57.5\% & 80.3\% \\
RFDNet                       & 580M  & 7.1MB  & 64.4\% & 85.8\% \\
\bottomrule
\end{tabular}
\label{imagenet}
\end{table}

\begin{table}[!t]
\renewcommand{\arraystretch}{1}
\centering
\caption{Detection results on PASCAL VOC 2007 dataset. The “07+12” means VOC07 trainval union with VOC12 trainval}
\begin{tabular}{lcccc}
\toprule
\multirow{2}{*}{{\begin{tabular}[l]{@{}c@{}} Model \end{tabular}}}  & \multirow{2}{*}{{\begin{tabular}[l]{@{}c@{}} Training \\ data \end{tabular}}}  & \multirow{2}{*}{{\begin{tabular}[c]{@{}c@{}} Input \\dimension \end{tabular}}}
& \multirow{2}{*}{{\begin{tabular}[c]{@{}c@{}} Model size\\(Parameters)\end{tabular}}} & \multirow{2}{*}{{\begin{tabular}[l]{@{}c@{}} mAP \end{tabular}}} \\
&&&& \\
\midrule
SqueezeNet-SSD  & 07+12  & 300$\times$300  & 21.1MB  & 64.3 \\
RFDNet-SSD      & 07+12  & 300$\times$300  & 17.2MB  & 70.1 \\
\bottomrule
\end{tabular}
\label{voc2007}
\end{table}

\begin{table}[!t]
\renewcommand{\arraystretch}{1}
\centering
\caption{Detection results on MS COCO dataset}
\setlength{\tabcolsep}{1.8mm}{
\begin{tabular}{lccccc}
\toprule
\multirow{2}{*}{{\begin{tabular}[l]{@{}c@{}} Model \end{tabular}}}     & \multirow{2}{*}{{\begin{tabular}[c]{@{}c@{}} Input \\dimension \end{tabular}}}
& \multirow{2}{*}{{\begin{tabular}[c]{@{}c@{}} Model size\\(Parameters)\end{tabular}}} & \multicolumn{3}{c}{Avg. Precision, IoU:} \\
&&&0.5:0.95&0.5&0.75 \\
\midrule
SqueezeNet-SSD    & 300$\times$300  & 55.4MB  & 8.4  & 15.2 & 8.2\\
RFDNet-SSD        & 300$\times$300  & 40.1MB  & 11.7 & 19.7 & 12.1\\
\bottomrule
\end{tabular}}
\label{coco2015}
\end{table}

\begin{table}[!t]
	\renewcommand{\arraystretch}{1}
	\centering
	\caption{Detection results of different DSFs on six datasets}
	\begin{tabular}{lcccc}
		\toprule
		Modules  & Width & mCDR/\%$\uparrow$      & mMDR/\%$\downarrow$     & mFDR/\%$\downarrow$ \\
		\midrule
		DSF9      &512$\times$1   & 98.09  & 1.26   & 0.65  \\
		DSF(4,9)    &256$\times$2   & 98.39  & 0.92   & 0.69  \\
		DSF(5,9)    &256$\times$2   & 94.13  & 4.27   & 1.60  \\
		DSF(6,9)    &256$\times$2   & 96.64  & 2.88   & 0.48  \\
		DSF(7,9)    &256$\times$2   & 97.90  & 1.64   & 1.46  \\
		DSF(8,9)    &256$\times$2   & 98.37  & 1.40   & 0.23  \\
		DSF(4,6,9)  &192$\times$3   & 97.99  & 1.64   & 0.37  \\
		DSF(4,7,9)  &192$\times$3   & 98.60  & 0.94   & 0.46  \\
		DSF(4,8,9)  &192$\times$3   & 98.51  & 0.86   & 0.63  \\
		DSF(5,6,9)  &192$\times$3   & 89.90  & 5.22   & 4.88  \\
		DSF(5,7,9)  &192$\times$3   & 95.15  & 3.44   & 1.41  \\
		\bottomrule
	\end{tabular}
	\label{diffmod}
\end{table}

\begin{table*}[!t]
	\renewcommand{\arraystretch}{1}
	\centering
	\caption{Detection results of connecting different modules, including SqueezeNet, RFDNet, MRPN, and MLPS}
	\setlength{\tabcolsep}{2.9mm}{
		\begin{tabular}{cccccccccccc}
			\toprule
			\multirow{2}{*}{SqueezeNet} & \multirow{2}{*}{RFDNet} & \multirow{2}{*}{MRPN} & \multirow{2}{*}{MLPS}    & \multirow{2}{*}{mCDR/\%$\uparrow$}  & \multirow{2}{*}{mMDR/\%$\downarrow$} & \multirow{2}{*}{mFDR/\%$\downarrow$} &
			\multirow{2}{*}{{\begin{tabular}[c]{@{}c@{}} Training \\ speed/s\end{tabular}}}   &
			\multirow{2}{*}{{\begin{tabular}[c]{@{}c@{}} Testing \\ speed/s\end{tabular}}}   &
			\multirow{2}{*}{{\begin{tabular}[c]{@{}c@{}} Memory \\ usage/MB\end{tabular}}}  &
			\multirow{2}{*}{{\begin{tabular}[c]{@{}c@{}} Model \\size/MB\end{tabular}}} \\
			& & & & & & & & & &  \\
			\midrule
			$\surd$ &--      &--      &--       & 97.12 & 1.15 & 1.73  & 0.085 & 0.026 & 745 & 20.7  \\
			--      &$\surd$ &--      &--       & 98.09 & 1.26 & 0.65  & 0.105 & 0.024 & 683 & 13.8  \\
			$\surd$ &--      &$\surd$ &--       & 97.90 & 0.91 & 1.19  & 0.114 & 0.027 & 770 & 21.6  \\
			--      &$\surd$ &$\surd$ &--       & 98.45 & 1.14 & 0.41  & 0.126 & 0.025 & 698 & 17.4  \\
			$\surd$ &--      &$\surd$ &$\surd$  & 98.36 & 0.70 & 0.94  & 0.118 & 0.028 & 795 & 25.1  \\
			--      &$\surd$ &$\surd$ &$\surd$  & 98.60 & 0.94 & 0.46  & 0.135 & 0.026 & 713 & 19.6  \\
			\bottomrule
	\end{tabular}}
	\label{modules}
\end{table*}

\begin{table*}[!t]
	\renewcommand{\arraystretch}{1}
	\caption{Detection results of six typical faults in comparison with state-of-the-art methods}
	\centering
	\setlength{\tabcolsep}{3.5mm}{
		%\resizebox{\textwidth}{!}{
		\begin{tabular}{lccccccccccccccc}
			\toprule
			\multirow{2}{*}{Methods}    & \multirow{2}{*}{mCDR/\%$\uparrow$}  & \multirow{2}{*}{mMDR/\%$\downarrow$} & \multirow{2}{*}{mFDR/\%$\downarrow$} &
			\multirow{2}{*}{{\begin{tabular}[c]{@{}c@{}} Training \\ speed/s\end{tabular}}}   &
			\multirow{2}{*}{{\begin{tabular}[c]{@{}c@{}} Testing \\ speed/s\end{tabular}}}   &
			\multirow{2}{*}{{\begin{tabular}[c]{@{}c@{}} Batch \\ size\end{tabular}}} &
            \multirow{2}{*}{{\begin{tabular}[c]{@{}c@{}} Model\\size/MB\end{tabular}}}&
            \multicolumn{2}{c}{\multirow{2}{*}{{\begin{tabular}[c]{@{}c@{}} Memory \\ usage/MB\end{tabular}}}} \\
			& & & & & & & &\\
			\midrule
			Cascade detector(LBP)       & 87.55   & 6.33   & 6.12   & --    & 0.048 & --  & 0.12  & \multicolumn{2}{c}{--}   \\
			HOG+Adaboost+SVM            & 93.32   & 3.25   & 3.43   & --    & 0.049 & --  & 0.11  & \multicolumn{2}{c}{--}   \\
			FAMRF+EHF                   & 94.96   & 1.00   & 4.04   & --    & 0.725 & --  & --    & \multicolumn{2}{c}{--}   \\
			\midrule
			SSD(VGG16)                  & 96.32   & 0.88   & 2.80   & 0.747 & 0.047 & 16  & 95.5  & \multicolumn{2}{c}{1173} \\
			YOLOv3                      & 88.85   & 2.58   & 8.57   & 3.537 & 0.026 & 64  & 246.3 & \multicolumn{2}{c}{1501} \\
			RefineDet(VGG16)            & 96.06   & 0.74   & 3.20   & 1.742 & 0.056 & 16  & 135.8 & \multicolumn{2}{c}{1415} \\
			RON(VGG16)                  & 98.15   & 0.47   & 1.38   & 0.892 & 0.029 & 32  & 157.9 & \multicolumn{2}{c}{1143} \\
			DSOD(DenseNet)              & 95.62   & 2.13   & 2.25   & 0.517 & 0.109 & 2   & 50.8  & \multicolumn{2}{c}{4429} \\
			\midrule
			MLKP(VGG16)                 & 98.21   & 0.68   & 1.11   & 0.722 & 0.147 & 128 & 596.1 & \multicolumn{2}{c}{3711} \\
			Faster R-CNN(VGG16)         & 98.19   & 0.96   & 0.85   & 0.289 & 0.065 & 128 & 546.8 & \multicolumn{2}{c}{1817} \\
			%CoupleNet(ResNet101)        & 97.51   & 0.22   & 2.27   & 0.851 & 0.112 & 128 & 409.6 & \multicolumn{2}{c}{3443} \\
			R-FCN(ResNet101)            & 94.68   & 1.71   & 3.61   & 0.524 & 0.096 & 128 & 199.9 & \multicolumn{2}{c}{3114} \\
			FTI-FDet(VGG16)             & 99.41   & 0.37   & 0.22   & 0.336 & 0.071 & 128 & 557.3 & \multicolumn{2}{c}{1823} \\
			Light FTI-FDet(VGG16)       & 99.22   & 0.32   & 0.46   & 0.318 & 0.058 & 128 & 89.7  & \multicolumn{2}{c}{1533} \\
            Cascade R-CNN(ResNet101)    & 97.34   & 0.96   & 1.70   & 0.615 & 0.203 & 2   & 220.8 & \multicolumn{2}{c}{3818} \\
			\midrule
			MobileNetV2-SSD             & 97.97   & 0.58   & 1.45   & 0.561 & 0.034 & 8   & 15.2  & \multicolumn{2}{c}{1343} \\
			MobileNetV2-SSDLite         & 94.65   & 0.29   & 5.06   & 0.101 & 0.018 & 16  & 12.3  & \multicolumn{2}{c}{827}  \\
			ShuffleNetV2-SSD            & 96.24   & 0.51   & 3.25   & 0.254 & 0.028 & 16  & 11.8  & \multicolumn{2}{c}{850}  \\
			Tiny-DSOD                   & 95.74   & 0.31   & 3.95   & 0.467 & 0.057 & 4   & 3.5   & \multicolumn{2}{c}{1469} \\
			Pelee(PeleeNet)             & 96.34   & 0.92   & 2.74   & 0.757 & 0.051 & 16  & 20.2  & \multicolumn{2}{c}{1412} \\
			\midrule
            Light FTI-FDet(RFDNet)      & 98.17   & 0.94   & 0.89   & 0.178 & 0.034 & 128 & 27.8  & \multicolumn{2}{c}{857}  \\
			RFDNet-SSD                  & 97.98   & 0.32   & 1.70   & 0.809 & 0.036 & 24  & 11.4  & \multicolumn{2}{c}{905}  \\
			LR FTI-FDet(RFDNet)         & 98.60   & 0.94   & 0.46   & 0.135 & 0.026 & 256 & 19.6  & \multicolumn{2}{c}{713}  \\
			\bottomrule
	\end{tabular}}
	\label{meanaccuray}
\end{table*}

\subsubsection{Evaluation Metrics} There are seven indexes: correct detection rate (CDR), missing detection rate (MDR), false detection rate (FDR), training speed, testing speed, test memory usage and model size (parameters) to evaluate the effectiveness of fault detectors. The indexes of CDR, MDR, and FDR are all used to measure the accuracy of detectors, which are calculated based on the method directly from~\cite{8911418}. For example, there is a test set which contains $m$ fault images and $n$ normal (non-fault) images, through the work of the detector, $a$ images are detected as fault, among them $b$ images are detected by error, meanwhile, $c$ images are detected as normal, among them $d$ images are detected by error. In this case, the indexes will be defined as:
\begin{equation}
  CDR = \frac{a+c}{m+n},\ MDR = \frac{b}{m+n},\ FDR = \frac{d}{m+n}.
\end{equation}
The mean value of CDRs, MDRs, and FDRs are calculated as mCDR, mMDR, and mFDR respectively to indicate the accuracy of fault detection for different datasets. Both model size and accuracy report the impact of CNN architectural designs~\cite{iandola2016squeezenet} on fault detectors. Both memory usage and training/testing speed reflect the dependence of detectors on hardware. Especially, we use the computational time for each iteration in training and testing phase for each image as training and testing speed, respectively. Memory usage is collected from a detector's memory usage on a single GPU in the testing phase.

\subsection{Performance Analysis}
\label{analysisframework}

\subsubsection{Backbone}
To verify the effectiveness of our RFDNet, we give a detailed discussion on the performance of RFDNet in comparison with the baseline light-weight network SqueezeNet. It can be seen from Table~\ref{imagenet} that RFDNet achieves a baseline of 64.4\% top-1 and 85.8\% top-5 accuracy on ImageNet, which is 6.9\% and 5.5\% higher than SqueezeNet with 1.4$\times$ less computation at the same size. The proposed RFDNet can also be deployed as an effective base network in object detection. We then perform experiments on VOC 2007 and MS COCO for detailed analysis of our RFDNet based on the SSD. Detection accuracy is measured by mean Average Precision (mAP) with 300 input resolutions. The experimental results on VOC2007 test set are summarized in Table~\ref{voc2007}. Our RFDNet achieves 70.1\% mAP, and its accuracy is higher than that of SqueezeNet by 5.8\% at only 81.2\% of model size. Moreover, the results on COCO minival set are summarized in Table~\ref{coco2015}. Our proposed RFDNet achieves 19.7\%/12.1\% with 0.5/0.75 IoU, which outperforms the SqueezeNet with a large margin. We observe that our [0.5:0.95] result is 3.3\% higher than the SqueezeNet at 72.4\% of model size. This indicates that our predicted locations are more accurate than the SqueezeNet with lower computational cost.

\subsubsection{Multi-scale Feature Utilization}
An important property of our method is that it combines coarse-to-fine information across deep CNN models. As an example, we compare different Conv. feature maps on six datasets to illustrate the superiority of the proposed multi-scale feature utilization (MFF\_1 and MFF\_2). Table~\ref{diffmod} shows the detection performance for connecting different DSF modules. “DSF(4,9)" means connecting DSF4 and DSF9 in both multi-RPN and MLPS score maps. “192$\times$3" means that we apply the 192-d 1$\times$1 Conv. layer on each of three DSF modules, respectively. In Table~\ref{diffmod}, the combination of DSF4, DSF7, and DSF9 works the best. The results indicate that the multi-layer combination performs roughly better than a single layer, and further verify the effectiveness of low-to-high combination strategy.

\subsubsection{Different Modules}
We analyze RFDNet, multi-RPN, and MLPS score maps by conducting experiments on six datasets. With the aforementioned computer, we only change the configuration of modules for a fair comparison. In Table~\ref{modules}, our RFDNet has higher accuracy and less computation than SqueezeNet. The combination of three modules in our framework can achieve the best performance. The index of mCDR significantly improves from 97.12\% to 98.60\%, and the testing speed is 0.026s. The results reveal that both MRPN and MLPS score maps can improve detection performance with few redundant computations. These two modules are able to learn more effective and comprehensive features than a single DSF for distinguishing faults from complex backgrounds.

\subsection{Comparison with State-of-the-art Methods}
To illustrate the superiority of our method, we compare our framework called as Light-weight Real-time FTI-FDet (LR FTI-FDet) with traditional detectors (Cascade detector with local binary pattern (LBP)~\cite{Sun2018Railway}, FAMRF + EHF~\cite{Sun2018Railway}, histogram of oriented gradient (HOG) + Adaboost + SVM~\cite{DollarABP14}), one-stage detectors (YOLOv3~\cite{redmon2018yolov3}, SSD~\cite{LiuAESRFB16}, RefineDet~\cite{zhang2018single}, RON~\cite{kong2017ron}, DSOD), two-stage detectors (Faster R-CNN~\cite{RenHGS15}, MLKP~\cite{wang2018multi}, R-FCN~\cite{DaiLHS16}, Cascade R-CNN~\cite{cai2018cascade}, FTI-FDet~\cite{zhang2018}, Light FTI-FDet~\cite{8911418}), and light-weight detectors (MobileNetV2-SSD~\cite{sandler2018inverted}, MobileNetV2-SSDLite~\cite{sandler2018inverted}, ShuffleNetV2-SSD~\cite{ma2018shufflenet}, Tiny-DSOD~\cite{yuxi2018tinydsod}, Pelee~\cite{wang2018pelee}).  In addition, we compare RFDNet-SSD with all above methods to discuss the performance of RFDNet and depthwise separable Conv.-based networks (\textit{e.g.} MobileNetV2) on fault detection. Specially, the related parameters in each detector are tuned to the best performance. %CoupleNet\footnote{https://github.com/tshizys/CoupleNet}~\cite{zhu2017couplenet},

\begin{figure}[!t]
  \centering
  \includegraphics[width=3.3in]{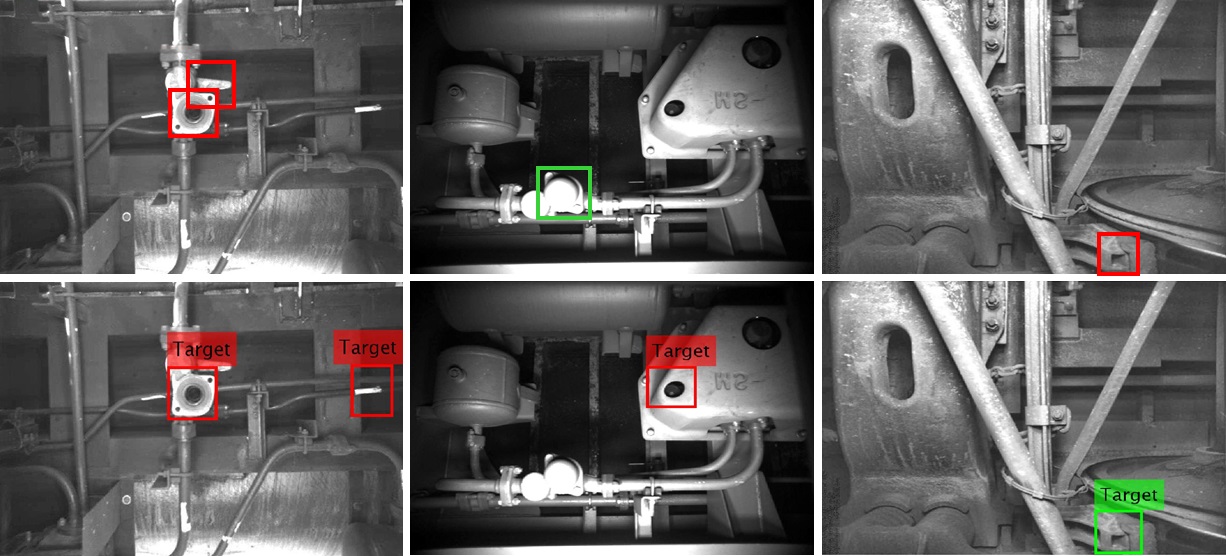}
  \caption{Visualization of ground-truths (top) and failure examples (bottom) obtained by our method. Green bounding boxes mean normal parts, and red bounding boxes are fault areas. Our method is unsatisfactory for the robustness of noise and the disturbance from other similar structures without faults.}
  \label{QFresults}
\end{figure}

\textbf{Accuracy and model size.}
As shown in Table~\ref{meanaccuray}, LR FTI-FDet achieves 98.60\% mCDR which outperforms RFDNet-SSD, all traditional methods, one-stage, light-weight and most two-stage detectors. The accuracy of both FTI-FDet and Light FTI-FDet are slightly higher than our method, but their model size is too large. Although the model size of each traditional methods is the smallest, but their accuracy is the lowest. The model sizes of our RFDNet-SSD and LR FTI-FDet are 11.4MB and 19.6 MB respectively, which is comparable to light-weight detectors and far less than all one- and two-stage detectors. After replacing backbone (VGG16) with RFDNet in Light FTI-FDet, our method achieves 0.43\% higher mCDR with 1.4$\times$ smaller than the Light FTI-FDet. Especially, the model size of LR FTI-FDet is 28.4/4.6$\times$ smaller than FTI-FDet/Light FTI-FDet with VGG16.
However, our method is unsatisfactory for the robustness of noise and the disturbance from other similar structures without faults. The comparisons between ground-truths and failure examples obtained by our method are shown in Fig.~\ref{QFresults}. We will solve it by expanding the datasets through adding more samples and performing data augmentation in the future. These operations will also improve the generalization ability of our method.

\textbf{Computational cost and speed.}
In Table~\ref{meanaccuray}, both training and testing speeds of our method are faster than traditional methods, one- and two-stage detectors. The testing speed ($>$38 fps) of our LR FTI-FDet is the same as YOLOv3 and 2.7/2.2$\times$ faster than FTI-FDet/Light FTI-FDet with 2.6/2.2$\times$ less memory usage. Our RFDNet-SSD has a comparable performance with MobileNetV2-SSD on fault detection while our method has smaller model size and memory usage. The speed of MobileNetV2-SSDLite is slight faster than our method, but its memory usage is higher. The main reason is that our LR FTI-FDet is a two-stage detector containing RPN and position-sensitive RoI pooling, which needs more computations than a one-stage detector MobileNetV2-SSDLite. But there are efficient DSF modules in RFDNet and many shared layers among RFDNet, multi-RPN, and MLPS score maps, so that the memory usage of our LR FTI-FDet is smaller, which merely needs 713 MB.

The experimental results confirm that our method achieves a much better trade-off between resources and accuracy than the state-of-the-art methods. The experiments on six typical fault datasets also indicate that our method is robust to the illumination variation with high versatility. Therefore, our method is the most suitable for real-time fault detection of freight train images, even though under strict memory and computational budget constraints.

%\begin{figure*}[!t]
%  \centering
%  \includegraphics[width=6.9in]{Figures/Figure4.jpg}
%  \caption{Qualitative results of our method.}
%  \label{QCresults}
%\end{figure*}

\section{Conclusion and future work}
\label{conclusion}
In this paper, we present a light-weight framework LR FTI-FDet in an end-to-end manner for real-time fault detection of freight train images in the wild. The proposed framework consists of a multi-RPN over RFDNet for fault proposal generation and MLPS score maps for fault proposal detection. Experiments show that the Top-1 accuracy of our RFDNet is 6.9\% higher than SqueezeNet with 1.4$\times$ less computation on ImageNet. Our RFDNet achieves 5.8\% mAP higher than SqueezeNet on VOC 2007, and our [0.5:0.95] result is 3.3\% higher than SqueezeNet on MS COCO. The detection results on six fault datasets indicate that our method is much faster during both training and testing as the light-weight detectors. Our method achieves competitive accuracy, 28.4/4.6$\times$ smaller model size and 2.6/2.2$\times$ less memory usage than FTI-FDet/Light FTI-FDet. The proposed LR FTI-FDet has lower resource requirements with the same testing speed as YOLOv3 up to 38 fps, 2.7/2.2$\times$ faster than FTI-FDet/Light FTI-FDet.

In the future, we plan to apply our method on embedded platforms (Raspberry Pi and Jetson Nano) to achieve real-time multi-fault detection in the wild, and further enhance accuracy and detection speed. 

\bibliographystyle{IEEEtran}
\bibliography{References}

%\section*{Acknowledgment}

\ifCLASSOPTIONcaptionsoff
  \newpage
\fi

\end{document}